# TITLE PAGE

**Title of the article:**
Cohort Characteristics and Factors Associated with Cannabis Use among Adolescents in Canada Using Pattern Discovery and Disentanglement Method

**Corresponding Author:**
Peiyuan Zhou, System Design Engineering, University of Waterloo, Waterloo, ON, Canada, email: p44zhou@uwaterloo.ca.

**Authors:**
Peiyuan Zhou[1], Andrew K.C. Wong[1], Yang Yang[2], Scott T. Leatherdale[2], Kate Battista[2], Zahid A. Butt[2], George Michalopoulos[2], Helen Chen[2]

[1]System Design Engineering, University of Waterloo, Waterloo, ON, Canada
[2]School of Public Health and Health Systems, University of Waterloo, Waterloo, ON, Canada

**Abstract**

**Background**

COMPASS is a longitudinal, prospective cohort study collecting data annually from students attending high school in jurisdictions across Canada. We aimed to discover significant frequent/rare associations of behavioral factors among Canadian adolescents related to cannabis use.

**Methods**

We use a subset of COMPASS dataset which contains 18,761 records of students in grades 9 to 12 with 31 selected features (attributes) involving various characteristics, from living habits to academic performance. We then used the Pattern Discovery and Disentanglement (PDD) algorithm that we have developed to detect strong and rare (yet statistically significant) associations from the dataset.

**Results**

PDD used the criteria derived from disentangled statistical spaces (known as Re-projected Adjusted-Standardized Residual Vector Spaces, notated as RARV). It outperformed methods using other criteria (i.e. *support* and *confidence*) popular as reported in literature. Association results showed that PDD can discover: i) a smaller set of succinct significant associations in clusters; ii) frequent and rare, yet significant, patterns supported by population health relevant study; iii) patterns from a dataset with extremely imbalanced groups (majority class: minority class = 88.3%: 11.7%).

**Conclusions**

Randomized controlled trial results on the COMPASS dataset have validated PDD's efficacy in discovering succinct interpretable frequent associations with comprehensive coverage and rare yet significant associations from datasets with extremely imbalanced class distribution without relying on any balancing process. The frequent associations show consistent results with common sense and literature surveys, while the rare patterns show very special cases. The success of PDD on this project indicates that PDD has great potential for population health data analysis.

**BACKGROUND AND SIGNIFICANCE**

Cannabis is considered as one of the most commonly used substances among young Canadians [1] [2] according to several nationally representative surveys conducted in the Canadian population [3] [4] [5] [6]. The Canadian Tobacco, Alcohol and Drugs Survey (CTADS) and the Canadian Student Tobacco, Alcohol and Drugs Survey (CSTADS) are two national surveys initiated by Health Canada for the prevalence of cannabis use among Canadians [3]. CTADS 2017 indicates that 19% of young Canadians aged 15 to 19 years have used cannabis in the past year [3]. CSTADS 2016/17 reports a prevalence of 17% among grades 7-12 students who have used cannabis in the past year [3]. According to data from the Canadian Cannabis Survey (CCS) 2019, 25% of the respondents aged 16 years and over have used cannabis in the past 12 months, an increase of 3% from the previous survey cycle [3]. Among the students who used cannabis more frequently in 2015, 72% of them reported having used cannabis in the past 90 days, while 34% had used cannabis on a weekly basis [4] [5]. Literature shows that over time, cannabis use among Canadian adolescents has increased. According to the 2010 Health Behaviour in School-Aged Children (HBSC) survey, about 40% of school-aged males and 37% of school-aged females had reported ever trying cannabis, a rate which had increased about 14% in both genders since 1990 [6].

Studies have shown that increased frequency of cannabis use among youth is generally associated with worse health outcomes and increased risk of depression, anxiety, and other mental health issues [7] [8]. Other health behaviours associated with increased frequency of cannabis use among youth include a higher likelihood of reporting cigarette smoking and binge drinking [7]. When looking at the transition of cannabis use patterns across time, it has been observed that in a cohort of Canadian high school students, about two-thirds of them either maintained or increased their cannabis use behaviors, whereas only one-third reduced usage or ceased to use [8].

Hence, to detect the associations of behavioral factors of students (taken as attribute value (AV), defined in Table 2), machine learning (ML) methods such as Decision Trees/Forests, Frequent Pattern Mining [9] [10] or Pattern Discovery (PD) [11] were explored. However, a serious limitation of these methods is that they typically produce overwhelming numbers of overlapping/redundant patterns coming from entwined groups/classes [12]. They are unable to partition and summarize the discovered patterns for revealing precise "knowledge" inherent in the data source environment, thus making interpretation difficult and lowering their prediction accuracy [13] [14]. For example, our bioinformatics studies [15] [16] [17] have provided strong scientific evidence that entangled associations exist in complex protein binding processes but can be disentangled by our novel algorithm [16] to reveal specific amino acid interaction functional spaces.

Along this line of research, we have developed a data-driven exploratory explainable method known as Pattern Discovery and Disentanglement (PDD). It is able to discover robust/succinct attribute value associations (AVAs) which are referred to as patterns if they are statistically significant based upon a hypothesis testing applying to the relational datasets. Such patterns will render statistical support and provide implicit functional clues that can be used to explain the underlying phenomena to augment scientific exploration. Recently, more supporting evidence of the efficacy of PDD in bioinformatics and biomedical applications has been attained [18]. It shows that in addition to genomic applications, PDD can reveal interpretable symptomatic patterns of individual patients from clinical/histopathological data and relate the findings to etiological/pathological origin after association disentanglement. It output the results in summarized form as well as a comprehensive representational framework referred to as PDD Knowledge Base (PDDKB) [18].

To extend the existing algorithm, PDD is enhanced to discover rare AVAs, which is a challenging topic in the current ML models. Generally, a rare pattern is defined as a pattern or an association with low frequency of occurrences. However, among the rare events discovered, the most difficult problem encountered is to distinguish whether it is an outlier, one which is not the target of our interest, or a significant rare case that may shed new light on the problem. This is the dilemma prevailing in ML. In this paper, we report the recent success of PDD in detecting both significant frequent as well as rare associations with cannabis use among adolescents in Canada.

## MATERIALS AND METHODS

### Data description

COMPASS [19] is a longitudinal, prospective cohort study that annually collects data from students attending secondary schools in jurisdictions across Canada. In the COMPASS study, student participants are asked about a variety of their personal health behaviors, including substance use, living habits, physical activity, and mental health information. In this study, we use a subset of the dataset collected through COMPASS in the school year of 2018-2019.

| Categorical of Attributes | Attribute Name | Data type | Attribute Description | Attribute Values |
|---|---|---|---|---|
| Target | Cannabis | Ordinal | last 12 months - how often use marijuana or cannabis | 1 = "I have never used marijuana"; 2 = "I have used marijuana but not in the last 12 months"; 3 = "Less than once a month"; 4 = "Once a month"; 5 = "2 or 3 times a month"; 6 = "Once a week"; 7 = "2 or 3 times a week"; 8 = "4 to 6 times a week"; 9 = "Every day"; |
| Substance Use | E-cigarette | Ordinal | last 30 days - how often use e-cigarette | 1 = "None"; 2 = "1 day"; 3 = "2 to 3 days"; 4 = "4 to 5 days"; 5 = "6 to 10 days"; 6 = "11 to 20 days"; 7 = "21 to 29 days"; 8 = "30 days (every day)" |
| | Drink | Ordinal | last 12 months - have 5 drinks | 1 = "I have never done this"; 2 = "I did not have 5 or more drinks on one occasion in the last 12 months"; 3 = "Less than once a month"; 4 = "Once a month"; 5 = "2 to 3 times a month"; 6 = "Once a week"; 7 = "2 to 5 times a week"; 8 = "Daily or almost daily" |
| | Smoking | Ordinal | last 30 days - smoking | 1 = "None"; 2 = "1 day"; 3 = "2 to 3 days"; 4 = "4 to 5 days"; 5 = "6 to 10 days"; 6 = "11 to 20 days"; 7 = "21 to 29 days"; 8 = "30 days (every day)" |
| Living Habit | Transport_To | Categorical | Travel To School | Car(Pass)  Car(Driver)  School Bus  Public Trans  Walk  Bicycle  Other |
| | Transport_From | Categorical | Travel From School | Car(Pass)  Car(Driver)  School Bus  Public Trans  Walk  Bicycle  Other |
| | Gambling | Boolean | last 30 days - gambling online | Yes  No |
| | PA_Hard | Boolean | >60 minutes of physical activity per day | Yes  No |
| | Smoke_Friend | Ordinal | closest friends smoke cigarettes | 0 = "None"; 1 = "1 friend"; 2 = "2 friends"; 3 = "3 friends"; 4 = "4 friends"; 5 = "5 or more friends" |
| | Skip_Class (H) | Ordinal | miss classes because of your health | 1 = "0 days"; 2 = "1 or 2 days"; 3 = "3 to 5 days"; 4 = "6 to 10 days"; 5 = "11 or more days" |
| | Sedentary | Numerical | Total daily sedentary | min = 0   max = 3550 |
| | PA_Total | Numerical | Total combined HARD and MODERATE physical activity | min = 0   max = 3990 |
| | Sleeping | Numerical | Sleeping : Minutes per day | min = 0   max = 585 |
| Personal Information | Race | Categorical | Race | White  Other  Asian  Black  Latin  Aboriginal |
| | BMI | Categorical | BMI (Categorical Values) | Under weight  Healthy  Over weight  Obese  NotState |
| | Province | Categorical | Province | ON  AB  BC  QC |
| | Grade | Categorical | Grade | Grade9  Grade11  Grade10  Grade12 |
| Academic Performance | Math_Score | Ordinal | Math Scores | 1 = "90% - 100%"; 2 = "80% - 89%"; 3 = "70% - 79%"; 4 = "60% - 69%"; 5 = "55% - 59%"; 6 = "50% - 54%"; 7 = "Less than 50%" |
| | Eng_Score | Ordinal | Enligh/French Scores | 1 = "90% - 100%"; 2 = "80% - 89%"; 3 = "70% - 79%"; 4 = "60% - 69%"; 5 = "55% - 59%"; 6 = "50% - 54%"; 7 = "Less than 50%" |
| | Skip_Class | Ordinal | last 4 weeks - skip classes | 1 = "0 classes"; 2 = "1 or 2 classes"; 3 = "3 to 5 classes"; 4 = "6 to 10 classes"; 5 = "11 to 20 classes"; 6 = "More than 20 classes" |
| | Education_Like | Categorical | the highest level of education - you would like to get | Master  Bachelor  DN  College  High School  <High School |
| | Education_Think | Categorical | the highest level of education - you think you will get | Master  Bachelor  DN  College  High School  <High School |
| School Environmnet | Support_PA | Categorical | School Support - physically active | VerySup  Support  USupport  VeryUSup |
| | Support_Health | Categorical | School Support - healthy foods and drinks | VerySup  Support  USupport  VeryUSup |
| | Support_Bully | Categorical | School Support -  no bullied | VerySup  Support  USupport  VeryUSup |
| | Support_Tobacco | Categorical | School Support - resist or quit tobacco | VerySup  Support  USupport  VeryUSup |
| | Support_drug_alc | Categorical | School Support - resist or quit drugs/alcohol | VerySup  Support  USupport  VeryUSup |
| Mental Health | CESD | Numerical | Center for Epidemiologic Studies Depression 10-Item Scale – Revised | min = 0   max = 30 |
| | Flourish | Numerical | Diener's Flourishing Scale | min = 8   max = 40 |
| | DERS | Numerical | Difficulties in Emotion Regulation Scale | min = 6   max = 30 |
| | GAD7 | Numerical | Generalized Anxiety Disorder 7-item Scale | min = 0   max = 21 |

**Figure 1. Description of the Compass Dataset.** The first column states the type of attributes, from living habits to mental health; the second column gives the name of the attribute; the third column shows the type of attribute values since the dataset is mixed-mode; the fourth column states the meanings of the attributes, and the last column shows the values assigned to the categorical attributes or the minimum value and maximum value of the numerical attributes. The data type (mode) of the attribute is shaded in different colors --- red for ordinal, green for the numeral, and blue for categorical.

The original dataset contains 74,501 records with 163 variables collected for all students. (Researchers interested in accessing the data can submit a data use request to the Principal Investigator, Dr. Leatherdale, using the application form found here: https://uwaterloo.ca/compass-system/information-researchers). The subset used in this study includes only students in grade 9 to 12 with no missing data resulting in a final sample of 18,761 records. The current analysis includes 31 attributes that reflect behavioral and personal characteristics of the students, from lifestyle choice to academic performance.

*Data preprocessing*

In the COMPASS dataset, a record is a row containing information of an individual student, and each column is associated with an attribute of categorical, Boolean, ordinal or numerical values. As for categorical and Boolean values, associations are represented directly by the association of the attribute values. However, for the attributes with numerical or ordinal values, to provide a robust definition of association and effectuate pattern interpretation, their numeral values are discretized into intervals which are then treated as categorical values. Figure 1 shows the types of data elements in our study dataset. The types of data include nominal (categorical, Boolean), and ordinal (numerical)) as highlighted in different colors (red for ordinal, green for the numeral, and blue for the categorical).

As Figure 1 shows, the ordinal attributes are labeled from 1 to 9 according to the ordinal order and its implied meaning. For example, the first row shows the target attribute with the name *"Cannabis"* which represents the frequency of usage in expressive terms used in the query such as: "In the last 12 months, how often did you use marijuana or cannabis? (a joint, pot, weed, hash)". The original values of the attributes and their distributions are shown in Table 1. According to their clinical meaning, we separated them into three levels as "Non-user" (contains labels 1-3), "Current" (contains labels 4-7), and "Regular" (contains labels 8,9). For all the other ordinal attributes, we discretize them into level values and label them by their meaning. In Figure 2, the original attributes as well as their discretization results are highlighted in red shade. The values under the column of "Attribute Values" are the discretization results and the values in the brackets show the original values.

As for the numerical attributes, such as the attributes representing the scores of different mental health indicators, we discretize their numerical values into interval values using the Equal Frequency discretization approach [20] which selected in a way that all bins contain approximately the same number of numerical values. The discretization results are also shown in Figure 2 and highlighted in the green shade.

| Data type of Attributes | Attribute Name | Attribute Description | Attribute Values Discretized Label (original value/[min,max]) | | | | |
|---|---|---|---|---|---|---|---|
| Ordinal | Cannabis | last 12 months - how often use cannabis | Non-user (1-3) | Occasional (4-7) | Regular (8,9) | | |
| | E-cigarette | last 30 days - how often use e-cigarette | Never (1) | [1 5] (1 to 5 days) | [6 29] (6 to 29 days) | Daily (everyday) | |
| | Drink | last 12 months - have 5 drinks | Barely (1-3) | Medium (4-7) | Daily (8) | | |
| | Smoking | last 30 days - smoking | Never (1) | Medium (2-7) | Daily (8) | | |
| | Smoke_Friend | closest friends smoke cigarettes | No (0) | [1 2] (1,2) | >2 (3,4,5) | | |
| | Skip_Class (H) | miss classes because of health | No (1) | [1 5] (2,3) | >5 (4,5) | | |
| | Math_Score | Math Scores | 80%+ (1,2) | 60%+ (3,4) | <60% (5,6,7) | | |
| | Eng_Score | Enligh/French Scores | 80%+ (1,2) | 60%+ (3,4) | <60% (5,6,7) | | |
| | Skip_Class | last 4 weeks - skip classes | No (1) | [1 5] (2,3) | >5 (4,5) | | |
| | | | Min | Max | After Discretization | | |
| Numerical | Sedentary | Total daily sedentary | 0 | 3550 | Short ([0,419]) | Medium ([420,644]) | Long ([645,3510]) |
| | PA_Total | Total combined HARD and MODERATE physical activity | 0 | 3990 | Low ([0,434]) | Medium ([435,870]) | High ([945 3990]) |
| | Sleeping | Sleeping : Minutes per day | 0 | 585 | Less ([0,389]) | Medium ([390,479]) | More ([480, 585]) |
| | CESD | Center for Epidemiologic Studies Depression 10-Item Scale – Revised | 0 | 30 | Health ([0, 9]) | Depression ([10,30]) | |
| | Flourish | Diener's Flourishing Scale | 8 | 40 | Languishing ([8, 29]) | Average ([30, 34]) | Flourishing ([35, 40]) |
| | DERS | Difficulties in Emotion Regulation Scale | 6 | 30 | Lack ([6,11]) | Average ([12,16]) | Normal ([17, 30]) |
| | GAD7 | Generalized Anxiety Disorder 7-item Scale | 0 | 21 | Mild ([0, 9]) | Moderate ([10, 14]) | Anxiety ([15, 21]) |

**Figure 2. Discretization Result for Ordinal attributes and Numerical Attributes**

| | | |
|---|---|---|
| 1 = "I have never used marijuana" | 13368 (71.25%) | Non-user (87.41%) |
| 2 = "I have used marijuana but not in the last 12 months" | 725 (3.86%) | |
| 3 = "Less than once a month" | 2306 (12.29%) | |
| 4 = "Once a month" | 554 (2.95%) | Current (9.48%) |
| 5 = "2 or 3 times a month" | 643 (3.43%) | |
| 6 = "Once a week" | 268 (1.43%) | |
| 7 = "2 or 3 times a week" | 313 (1.67%) | |
| 8 = "4 to 6 times a week" | 256 (1.36%) | Regular (3.11%) |
| 9 = "Every day" | 328 (1.75%) | |

**Table 1. Original Value Distribution for the attribute *"Cannabis"***

As Table 1 shows, the COMPASS class distribution in the dataset is extremely imbalanced. It contains 87.41% records labeled as Non-user, but only 3.11% records as Regular. The number of records in the majority class is almost 30 times over that of the minority class. When datasets are imbalanced in diagnostic categories, the ordinary ML methods might produce results overwhelmed by the majority classes while diminishing significant patterns in minority classes [21]. To balance the dataset, researchers usually use under-sampling and oversampling methods [22]. However, such balancing strategies may meet one or more of the following limitations:

i) The current ML methods (i.e., association analysis) need to be guided by the class labels (target attribute),

ii) Oversampling may result in overfitting towards the minority class samples especially for higher over-sampling rates [22] [23].

iii) Under-sampling may result in deleting significant information from the majority class [24].

However, PDD has overcome all these problems. It discovers associations without using class information and sampling methods. It shows robust performance for discovering patterns from an imbalanced dataset [21]. No matter whether the data is balanced or imbalanced, PDD can naturally handle it robustly and steadily [21].

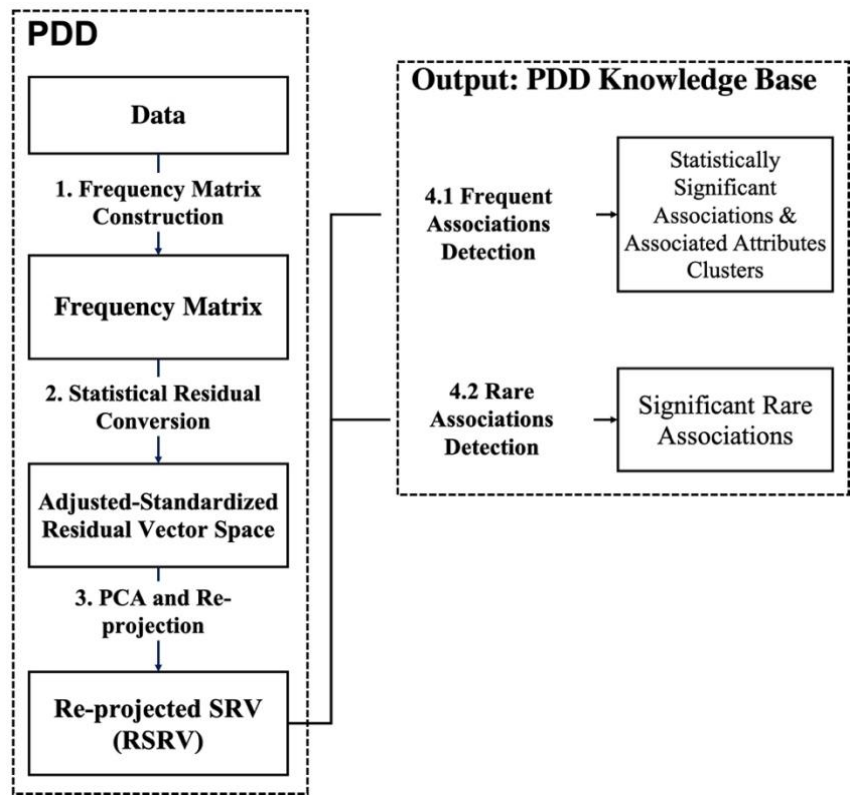

**Figure 3. Overview of PDD.**

*Methodology of Pattern discovery and disentanglement (PDD)*

After discretizing the numerical values into quantized intervals, each attribute would assume a finite set of discrete values. In a formal term, the input data has been transformed into a dataset containing attributes with distinct event values. Then, PDD is implemented as illustrated in Figure 3. Table 2 shows the terminologies used in the paper.

| Terminology | Definition |
|---|---|
| Attribute Value (AV) | A dataset contains **N** attributes (denoted as A = $\{A_1, A_2, ... A_N\}$, and each attribute ($A_n$) contains a set of attribute values. (e.g. the attribute "Cannabis" contains three attribute values: "Non-user", "Current" and "Regular") |
| Attribute Value Association (AVA) | An AVA refers to a pair of attribute values from two different attributes. If specially specified, it could refer to the association of a set of AVs. |
| Adjusted standard residual (AR) | AR for an AVA accounts for the deviation of the observed frequency of occurrences of the AVA against the expected frequency of its occurrences if the AVs in the AVA are statistically independent |
| AR Vector Space (ARV) | AR V is a matrix that contains all AR values for each distinct AV pair found in **R**. So, the row of the AR V corresponds to a vector of AR for a distinct AV. |
| Principal Components (PC) | One of the *n* Principal Components (PCs) obtained from Principal Component Analysis (PCA) when applied to the ARV. |
| Reprojected ARV (RARV) | An RARV is an ARV corresponding to a PC showing the transformed position (coordinates) of all the AV-vector projected onto the PC after PCA. |

**Table 2 Terminology.** The terminology table succinctly lists and briefly defines terminologies used in the paper.

To measure the association between an attribute-value (AV) pair, *Frequency* is the most popular approach to describe the relative frequencies between two attribute values. However, *Frequency* is difficult to measure how significant of the association. Hence, as reported in our previous paper [21], the adjusted standardized residual (AR) is more appropriate to represent the statistical weights of the AV pair than their frequency of co-occurrences. The AR for an AV pair (say. $AV_1$ and $AV_2$) is denoted as $AR(AV_1 \leftrightarrow AV_2)$ and calculated by Eqn. (1).

$$AR(AV_1 \leftrightarrow AV_2) = \frac{Occ(AV_1 \leftrightarrow AV_2) - Exp(AV_1 \leftrightarrow AV_2)}{\sqrt{Exp(AV_1 \leftrightarrow AV_2)}} * \left(1 - \frac{Occ(AV_1)}{M} \frac{Occ(AV_2)}{M}\right) \quad (1)$$

where $Occ(AV_1)$ and $Occ(AV_2)$ are the number of occurrences of AV; $Occ(AV_1 \leftrightarrow AV_2)$ is the total number of co-occurrence for AV pair; and $Exp(AV_1 \leftrightarrow AV_2)$ is the expected frequency and M is the total number of records.

So, after the first two steps, the original data is transformed into the AR Vector Space (ARV), which is a T x T matrix, where T represents the total number of distinct AVs for all attributes in the dataset. Each item in the matrix is the AR value of an AV pair, which accounts for the deviation of its observed frequency of occurrences against the expected frequency if the AVs in the pair are statistically independent. Generally speaking, the significant associations are selected according to the threshold obtained from the hypothesis test for statistically significant AR. For example, when the association's AR >1.96, it is treated as positively significant with a 95% confidence level (AR >1.44 with 85% and AR >1.28 with 80% respectively). A row in ARV corresponding to an AV is defined as the AV-vector where its i[th] coordinate is the AR of the AV pair of that AV associating with the AV corresponding to the i[th] column vector in the ARV.

Then, the Principal Component Analysis (PCA) is applied to decompose the ARV in several PCs each of which is reprojected to an ARV (with same basis vectors) referred to as Reprojected ARVs (RARVs). Each item in RARVs is denoted as Reprojected AR (RAR). Like the AR for an AV pair in the ARV (step 4.1), we say that the association of that AV pair in the RARV is significant if its RAR>1.96. Furthermore, PDD can generate Attribute Value Clusters (AV Clusters) from that small set of disentangled spaces (DS's) [25]. In each DS, usually, the number of clusters is small since most of the AV-vectors not contributing to eigenvalues of that PC are close to the center of the PC and usually contain insignificant RAR in the RARV. To reveal the joint association of the AVs in the cluster, the AVs that are grouped into the same cluster must have a significant AV association with other AVs in the same cluster.

We next extend step 4.1 to step 4.2 to discover rare patterns. In a general setting, a rare pattern is defined as an association with a low frequency of occurrences. For example, Apriori Inverse [26] algorithm finds a perfectly rare pattern if it is one having the *support* not lower than a minimum *support* threshold nor higher than a maximum *support* threshold. It emphasizes the rarity of the pattern but not its statistical significance. However, a rare pattern with low frequency may be caused by various reasons. For example, some low-frequency cases caused by the class imbalance in the dataset are not really rare patterns. In this case, the AR value would render a better criterion to evaluate the rare patterns than the frequency. Furthermore, a rare pattern may also be just an outlier which is not a significant rare case. Hence, to assess the statistical significance of a rare pattern, its RAR value in certain RARV where such association is spotted should be used instead. Thus, in PDD, both RAR and AR are used as the criteria for detecting rare patterns. As mentioned in Step 3, PCA is sensitive to the relative scaling of the original frequency values used. By employing RAR instead, PDD can produce succinctly statistically significant patterns even if they are rare. Hence, in this study, we define a significant rare pattern as one has high RAR above 1.96, AR lower than -1.96, and frequency is not 0. If we wish to detect even rarer AVAs, we could lower the RAR threshold as well.

**COMPARISON RESULTS**

In this study, we conducted experiments on the COMPASS dataset and compared the discovered associations using PDD and other criteria. Statistical criteria, such as *support*, *confidence* and *adjusted standard residual* (AR) are used for discovering associations assessment. For example, the well-known association analysis technique, Apriori [10] [27], uses *support* and *confidence* to measure how items (attribute values) are associated to each other. *Support* is used as the same as the *frequency*, denoted as $sup\ (AV_1, AV_2)$ which represents the proportion of a pair of attribute values ($AV_1\ and\ AV_2$) that appears in the data together. And *confidence* denotes how likely $AV_1$ occurs when $AV_2$ occurs. It is expressed as $con(AV_1 \rightarrow AV_2) = \frac{\sup(AV_1, AV_2)}{\sup(AV_1)}$. The value of both *support* and *confidence* is in the range of [0,1].

An association with higher *support* is more significant than the association with lower *support*, and the user has to set various thresholds themselves to identify the association with support values above the threshold as a significant association. However, the COMPASS dataset is extremely imbalanced, so *support* which describes the co-occurrence of attribute values is not effective since it only discovers the associations with the majority class (e.g., Cannabis = Non-user in this case). Compared to *support*, confidence measures by the proportion of records with AV1, in which AV2 also appears, but one drawback of the confidence measure is that it might misrepresent the importance of an association.

### Frequent Patterns: Characteristics Associated with Cannabis Use

| Cannabis Use = Non-user | | | | | Cannabis Use = Regular | | | | | Cannabis Use = Current | | | | |
|---|---|---|---|---|---|---|---|---|---|---|---|---|---|---|
| Group 1 (Discovered in DS1,PG1, SubPG1) | | | | | Group 1 (Discovered in DS1,PG2, SubPG1) | | | | | Group 1 (Discovered in DS1,PG2, SubPG3) | | | | |
| Support | Confidence | SR | Attribute | Attribute Value | Support | Confidence | SR | Attribute | Attribute Value | Support | Confidence | SR | Attribute | Attribute Value |
| 0.68 | 0.96 | 56.37 | E-Cigarette | Never | 0.01 | 0.39 | 38.71 | E-Cigarette | Daily | 0.05 | 0.55 | 44 | Drink | Medium |
| 0.77 | 0.93 | 53.05 | Drink | Barely | 0.01 | 0.34 | 30.19 | Smoking | Medium | 0.02 | 0.4 | 34.9 | Smoking | Medium |
| 0.85 | 0.97 | 52.08 | Smoking | Never | 0.01 | 0.31 | 27.78 | Smoke_Friend | >2 | 0.02 | 0.41 | 33.4 | E-Cigarette | Daily |
| 0.71 | 0.92 | 36.03 | Smoke_Friend | No | 0.02 | 0.54 | 23.68 | Drink | Medium | 0.03 | 0.31 | 31.9 | E-Cigarette | [6 29] |
| 0.63 | 0.93 | 32.21 | Skip_Class | No | 0.01 | 0.24 | 23.31 | Skip_Class | >5 | 0.05 | 0.49 | 20.5 | Skip_Class | [1 5] |
| 0.51 | 0.92 | 20.31 | Math_Score | 80%+ | 0.01 | 0.2 | 16.48 | Eng_Score | <60% | 0.03 | 0.35 | 20 | Smoke_Friend | [1 2] |
| 0.56 | 0.91 | 18.58 | Skip_Class(H) | No | 0.01 | 0.3 | 15.81 | Math_Score | <60% | 0.01 | 0.22 | 14.3 | Smoke_Friend | >2 |
| 0.53 | 0.91 | 17.54 | Eng_Score | 80%+ | 0.01 | 0.25 | 13.89 | E-Cigarette | [6 29] | 0.02 | 0.18 | 12.9 | Transport_To | Car(Driver) |
| 0.39 | 0.92 | 14.93 | Education_Like | >=Master | 0.01 | 0.35 | 12.95 | Support_Tobacco | Very Unsupport | 0.02 | 0.25 | 12.8 | E-Cigarette | [1 5] |
| 0.86 | 0.98 | 14.42 | Gambling | No | 0.01 | 0.21 | 12.32 | Support_Bully | Very Unsupport | 0.02 | 0.18 | 12.4 | Transport_From | Car(Driver) |
| 0.6 | 0.9 | 14.34 | CESD | health | 0.01 | 0.35 | 12.06 | Education_Like | College | 0.03 | 0.34 | 11.5 | DERS | normal |
| 0.22 | 0.93 | 13.79 | Grade | G9 | 0.01 | 0.36 | 11.5 | Smoke_Friend | [1 2] | 0.04 | 0.46 | 11.5 | CESD | depression |
| 0.13 | 0.95 | 12.15 | Race | Asian | 0.02 | 0.49 | 11.32 | Skip_Class | [1 5] | 0.05 | 0.48 | 11.5 | Skip_Class(H) | [1 5] |
| 0.4 | 0.91 | 11.98 | Sedentary | short | 0.02 | 0.54 | 10.8 | Flourish | languishing | 0.03 | 0.32 | 11.4 | Education_Think | College |
| 0.28 | 0.92 | 11.85 | Education_Think | >=Master | 0.01 | 0.32 | 10.66 | Support_Drug_Alc | Very Unsupport | 0.02 | 0.18 | 11.4 | Math_Score | <60% |
| 0.38 | 0.91 | 11.69 | DERS | lack | 0.01 | 0.36 | 10.31 | Grade | G12 | 0.03 | 0.35 | 11.3 | Sedentary | long |
| 0.69 | 0.89 | 11.55 | GAD7 | mild | 0.01 | 0.36 | 8.77 | Education_Think | College | 0.02 | 0.26 | 10.6 | Education_Like | College |
| 0.33 | 0.91 | 11.36 | Support_Tobacco | Support | 0.01 | 0.38 | 8.07 | Sedentary | long | 0.03 | 0.29 | 10.3 | Grade | G12 |
| 0.33 | 0.91 | 10.61 | Support_Drug_Alc | Support | 0.02 | 0.49 | 8 | CESD | depression | 0.02 | 0.24 | 10.3 | Support_Tobacco | Very Unsupport |
| 0.27 | 0.91 | 10.1 | Flourish | Flourishing | 0.01 | 0.36 | 7.84 | DERS | normal | 0.01 | 0.19 | 9.89 | Skip_Class | >5 |
| 0.46 | 0.89 | 8.83 | Support_Bully | Support | 0.01 | 0.19 | 7.74 | Transport_To | Car(Driver) | 0.02 | 0.24 | 9.78 | Support_Drug_Alc | Very Unsupport |
| 0.16 | 0.91 | 6.94 | Province | BC | 0.01 | 0.19 | 7.49 | Transport_From | Car(Driver) | 0.01 | 0.18 | 8.9 | Eng_Score | <60% |
| 0.52 | 0.89 | 6.83 | PA_Hard | No | 0.01 | 0.18 | 7.35 | GAD7 | anxiety | 0.04 | 0.43 | 8.78 | Math_Score | 60%+ |
| 0.16 | 0.91 | 6.76 | Support_Bully | Very Support | 0.01 | 0.47 | 6.21 | Skip_Class(H) | [1 5] | 0.04 | 0.39 | 8.57 | Sleeping | low |
| 0.35 | 0.89 | 6.25 | Transport_From | School bus | 0.02 | 0.53 | 5.86 | PA_Hard | Yes | 0.03 | 0.31 | 8.39 | Support_Bully | Unsupport |
| 0.35 | 0.89 | 6.13 | Support_PA | Very Support | 0.02 | 0.66 | 5.82 | Province | ON | 0.01 | 0.16 | 8.37 | Support_Bully | Very Unsupport |
| 0.17 | 0.9 | 5.84 | Sleeping | high | 0.01 | 0.47 | 5.75 | Eng_Score | 60%+ | 0.04 | 0.45 | 8.34 | Eng_Score | 60%+ |
| 0.2 | 0.9 | 5.8 | Province | QC | 0.01 | 0.41 | 5.7 | Sleeping | low | 0.04 | 0.41 | 7.85 | Flourish | languishing |
| 0.25 | 0.9 | 5.56 | Grade | G10 | 0.01 | 0.38 | 4.24 | PA_Total | high | 0.01 | 0.15 | 7.71 | GAD7 | anxiety |
| 0.12 | 0.91 | 5.18 | Support_Drug_Alc | Very Support | 0.01 | 0.41 | 3.51 | Math_Score | 60%+ | 0.01 | 0.18 | 7.65 | Skip_Class(H) | >5 |
| 0.46 | 0.89 | 5.12 | Sleeping | medium | 0.01 | 0.19 | 2.85 | E-Cigarette | [1 5] | 0.01 | 0.14 | 5.88 | Support_PA | Unsupport |
| 0.12 | 0.91 | 4.96 | Support_Tobacco | Very Support | 0.01 | 0.32 | 1.96 | Grade | G11 | 0.03 | 0.34 | 5.39 | Grade | G11 |
| 0.34 | 0.89 | 4.88 | Transport_To | School bus | | | | | | 0.02 | 0.17 | 4.7 | GAD7 | moderate |
| 0.17 | 0.89 | 4.05 | Transport_From | walking | | | | | | 0.01 | 0.14 | 4.65 | Support_Health | Very Unsupport |
| 0.47 | 0.88 | 3.9 | Support_Health | Support | | | | | | 0.04 | 0.39 | 4.39 | Support_Drug_Alc | Unsupport |
| 0.25 | 0.89 | 3.83 | Education_Think | Bachelor | | | | | | 0.04 | 0.46 | 4.26 | PA_Hard | Yes |
| 0.27 | 0.89 | 3.41 | Transport_To | Car(Passenger) | | | | | | 0.04 | 0.4 | 4.07 | Support_Tobacco | Unsupport |
| 0.33 | 0.88 | 2.73 | Flourish | average | | | | | | 0.01 | 0.13 | 3.14 | Education_Think | High School |
| 0.66 | 0.88 | 2.1 | BMI | Healthy | | | | | | 0.02 | 0.22 | 2.6 | Support_Health | Unsupport |
| Group 2 (Discovered in DS2,PG1, SubPG1) | | | | | Group 2 (Discovered in DS2,PG2, SubPG1) | | | | | Group 2 (Discovered in DS2,PG2, SubPG3) | | | | |
| Support | Confidence | SR | Attribute | Attribute Value | Support | Confidence | SR | Attribute | Attribute Value | Support | Confidence | SR | Attribute | Attribute Value |
| 0.68 | 0.96 | 56.37 | E-Cigarette | Never | 0.01 | 0.39 | 38.71 | E-Cigarette | Daily | 0.05 | 0.55 | 43.99 | Drink | Medium |
| 0.77 | 0.93 | 53.05 | Drink | Barely | 0.01 | 0.34 | 30.19 | Smoking | Medium | 0.02 | 0.4 | 34.94 | Smoking | Medium |
| 0.85 | 0.97 | 52.08 | Smoking | Never | 0.01 | 0.31 | 27.78 | Smoke_Friend | >2 | 0.02 | 0.41 | 33.38 | E-Cigarette | Daily |
| 0.71 | 0.92 | 36.03 | Smoke_Friend | No | 0.02 | 0.54 | 23.68 | Drink | Medium | 0.03 | 0.31 | 31.9 | E-Cigarette | [6 29] |
| 0.51 | 0.92 | 20.31 | Math_Score | 80%+ | 0.01 | 0.24 | 23.31 | Skip_Class | >5 | 0.03 | 0.35 | 20.02 | Smoke_Friend | [1 2] |
| 0.53 | 0.91 | 17.54 | Eng_Score | 80%+ | 0.01 | 0.2 | 16.48 | Eng_Score | <60% | 0.01 | 0.22 | 14.26 | Smoke_Friend | >2 |
| 0.39 | 0.92 | 14.93 | Education_Like | >=Master | 0.01 | 0.3 | 15.81 | Math_Score | <60% | 0.02 | 0.18 | 12.89 | Transport_To | Car(Driver) |
| 0.13 | 0.95 | 12.15 | Race | Asian | 0.01 | 0.25 | 13.89 | E-Cigarette | [6 29] | 0.02 | 0.18 | 12.44 | Transport_From | Car(Driver) |
| 0.16 | 0.91 | 6.94 | Province | BC | 0.01 | 0.35 | 12.06 | Education_Like | College | 0.03 | 0.32 | 11.42 | Education_Think | College |
| 0.52 | 0.89 | 6.83 | PA_Hard | No | 0.01 | 0.36 | 11.5 | Smoke_Friend | [1 2] | 0.02 | 0.18 | 11.41 | Math_Score | <60% |
| 0.17 | 0.89 | 4.05 | Transport_From | walking | 0.02 | 0.49 | 11.32 | Skip_Class | [1 5] | 0.02 | 0.26 | 10.58 | Education_Like | College |
| 0.47 | 0.88 | 3.9 | Support_Health | Support | 0.01 | 0.36 | 8.77 | Education_Think | College | 0.01 | 0.19 | 9.89 | Skip_Class | >5 |
| 0.25 | 0.89 | 3.83 | Education_Think | Bachelor | 0.01 | 0.38 | 8.07 | Sedentary | long | 0.01 | 0.18 | 8.9 | Eng_Score | <60% |
| 0.27 | 0.89 | 3.41 | Transport_To | Car(Passenger) | 0.01 | 0.19 | 7.74 | Transport_To | Car(Driver) | 0.04 | 0.43 | 8.78 | Math_Score | 60%+ |
| 0.12 | 0.89 | 2.24 | Transport_To | walking | 0.01 | 0.19 | 7.49 | Transport_From | Car(Driver) | 0.04 | 0.45 | 8.34 | Eng_Score | 60%+ |
| 0.2 | 0.88 | 2.16 | PA_Total | low | 0.02 | 0.53 | 5.86 | PA_Hard | Yes | 0.06 | 0.62 | 6.95 | Province | ON |
| | | | | | 0.02 | 0.66 | 5.82 | Province | ON | 0.04 | 0.46 | 4.26 | PA_Hard | Yes |
| | | | | | 0.01 | 0.47 | 5.75 | Eng_Score | 60%+ | 0.07 | 0.73 | 3.98 | Race | White |
| | | | | | 0.01 | 0.38 | 4.24 | PA_Total | high | 0.01 | 0.13 | 3.14 | Education_Think | High School |
| | | | | | 0.01 | 0.41 | 3.51 | Math_Score | 60%+ | 0.03 | 0.33 | 2.02 | PA_Total | high |
| | | | | | 0.01 | 0.19 | 2.85 | E-Cigarette | [1 5] | | | | | |
| | | | | | Group 3 (Discovered in DS3,PG1, SubPG1) | | | | | Group 3 (Discovered in DS3,PG2, SubPG1) | | | | |
| | | | | | Support | Confidence | SR | Attribute | Attribute Value | Support | Confidence | SR | Attribute | Attribute Value |
| | | | | | 0.01 | 0.35 | 12.95 | Support_Tobacco | Very Unsupport | 0.05 | 0.55 | 43.99 | Drink | Medium |
| | | | | | 0.01 | 0.21 | 12.32 | Support_Bully | Very Unsupport | 0.02 | 0.4 | 34.94 | Smoking | Medium |
| | | | | | 0.02 | 0.54 | 10.8 | Flourish | languishing | 0.03 | 0.32 | 11.42 | Education_Think | College |
| | | | | | 0.01 | 0.32 | 10.66 | Support_Drug_Alc | Very Unsupport | 0.02 | 0.26 | 10.58 | Education_Like | College |
| | | | | | 0.01 | 0.36 | 10.31 | Grade | G12 | 0.04 | 0.43 | 8.78 | Math_Score | 60%+ |
| | | | | | 0.01 | 0.38 | 8.07 | Sedentary | long | 0.04 | 0.45 | 8.34 | Eng_Score | 60%+ |
| | | | | | 0.02 | 0.49 | 8 | CESD | depression | 0.04 | 0.39 | 4.39 | Support_Drug_Alc | Unsupport |
| | | | | | 0.01 | 0.36 | 7.84 | DERS | normal | 0.04 | 0.4 | 4.07 | Support_Tobacco | Unsupport |
| | | | | | 0.01 | 0.18 | 7.35 | GAD7 | anxiety | 0.07 | 0.73 | 3.98 | Race | White |
| | | | | | 0.02 | 0.53 | 5.86 | PA_Hard | Yes | | | | | |
| | | | | | 0.02 | 0.66 | 5.82 | Province | ON | | | | | |
| | | | | | 0.01 | 0.41 | 5.7 | Sleeping | low | | | | | |
| | | | | | 0.01 | 0.38 | 4.24 | PA_Total | high | | | | | |

**Table 3. Discovered Frequent Associations for Cannabis Use.**

Both *support* and *confidence* together are used for discovering association in Apriori [10], yet for some associations within the same range, it is difficult to determine whether the association is significant enough. Therefore, AR is the best choice since it not only gives an adjustable statistical threshold = 0 to indicate positive or negative patterns but also gives a significance threshold, like 1.96 for 95% confidence level, to reveal their statistical significance and rejects the random associations in between, which means, when AR>1.96, the association is positively significant (the association more likely occur), and when AR<-1.96, the association is negatively significant (the association more unlikely occur). Furthermore, although ARV can detect the statistically significant associations, yet the statistical associations it captures could be entangled with others as reported in our previous studies [17] [28] [29]. Therefore, the RAR criterion is much more meaningful and superior to the frequency and the lift in revealing significant attribute value associations.

Finally, to summarize the attribute-value associations discovered from PDD, Table 3 and Table 4 presents respectively the significant frequent associations (RAR>1.96 and AR>1.96) and significant rare associations (RAR>1.96 and AR<-1.96) between target values and other attribute values in hierarchical clusters (groups).

As Table 3 shows, we present the attribute values significantly associated with Cannabis. The three main columns show associations for Cannabis = "Non-user", Cannabis = "Regular", and Cannabis = "Current" respectively. There are two groups of attribute values, discovered from the first two disentangled spaces in Pattern Group 1 (PG1), which are significantly associated with Cannabis = "Non-user". It should be noted, that the records labeled as "Non-user" have the largest proportion in the dataset, so their associations are discovered first. Then attribute value clusters associated with Cannabis = "Regular" and Cannabis = "Current" are also discovered in the first two disentangled spaces. In addition, Cannabis = "Non-user" is also discovered in Pattern Group 2 (PG2). Finally, in the third disentangled space (DS3), the minority groups, Cannabis = "Regular" and Cannabis = "Current", are succinctly unveiled in Pattern Group 1 (PG1) and Pattern Group 2 (PG2) respectively.

In addition, the association results obtained by PDD are consistent with those obtained from common sense and our literature survey. For example, "Current" and "Regular" cannabis use was associated with low English (Eng_Score = "<60%") and Math (Math_Score = "<60%") performance. On the contrary, the Non-user (Cannabis = "Non-user") associates with better English (Eng_Score = "80%+") and Math (Math_Score = "80%+") academic performance. This is consistent with other research [30] showing that cannabis use in adolescence is a negative effect on educational outcomes.

Furthermore, previous studies [31] [32] [33] have shown that cannabis use is strongly associated with mental health attributes, which are also presented in our result (Table 3). For example, as Table 3 shows, the score of the "Center for Epidemiologic Studies Depression Scale" (CESD) = "health" is associated with "Non-user"; and that CESD=depression is associated with Cannabis = "Regular". Similarly, Flourish = "flourishing" or "average" is associated with Cannabis = "Non-user" while Flourish = "languishing" is associated with Cannabis = "Regular" and Cannabis = "Current".

| Rare Patterns: Characteristics Associated with Cannabis Use | | | | | | | | | | | | | | |
|---|---|---|---|---|---|---|---|---|---|---|---|---|---|---|
| Cannabis Use = Non-user | | | | | Cannabis Use = Regular | | | | | Cannabis Use = Current | | | | |
| Group 1 (Discovered in DS2,PG1, SubPG1) | | | | | Group 1 (Discovered in DS2,PG2, SubPG1) | | | | | Group 1 (Discovered in DS2,PG2, SubPG3) | | | | |
| Support | Confidence | SR | Attribute | Attribute Value | Support | Confidence | SR | Attribute | Attribute Value | Support | Confidence | SR | Attribute | Attribute Value |
| 0.12 | 0.78 | -16 | Support_Tobacco | Very Unsupport | 0.01 | 0.38 | -11.8 | Skip_Class(H) | No | 0.05 | 0.54 | -11.5 | CESD | health |
| 0.28 | 0.83 | -14 | CESD | depression | 0.02 | 0.51 | -8 | CESD | health | 0.03 | 0.31 | -9.96 | DERS | lack |
| 0.18 | 0.81 | -14 | DERS | normal | 0.01 | 0.17 | -6.61 | Flourish | Flourishing | 0.07 | 0.69 | -9.17 | GAD7 | mild |
| 0.13 | 0.8 | -14 | Support_Drug_Alc | Very Unsupport | 0.02 | 0.66 | -6.59 | GAD7 | mild | 0.02 | 0.22 | -7.52 | Flourish | Flourishing |
| 0.06 | 0.76 | -14 | Support_Bully | Very Unsupport | 0.01 | 0.32 | -5.72 | Sedentary | short | 0.01 | 0.12 | -6.09 | Support_Bully | Very Support |
| 0.27 | 0.83 | -13 | Flourish | languishing | 0.01 | 0.3 | -5.55 | DERS | lack | 0.01 | 0.13 | -5.98 | Sleeping | high |
| 0.07 | 0.79 | -11 | GAD7 | anxiety | 0.01 | 0.3 | -4.81 | Transport_From | School bus | 0.03 | 0.33 | -5.44 | Support_PA | Very Support |
| 0.25 | 0.84 | -11 | Sleeping | low | 0.01 | 0.31 | -3.76 | Transport_To | School bus | 0.01 | 0.1 | -4.77 | Support_Drug_Alc | Very Support |
| 0.03 | 0.77 | -8.8 | Support_Health | Very Unsupport | 0.01 | 0.34 | -2.53 | Support_PA | Very Support | 0.01 | 0.09 | -4.4 | Support_Tobacco | Very Support |
| 0.19 | 0.84 | -7.9 | Support_Bully | Unsupport | 0.02 | 0.64 | -2.39 | Race | White | 0.03 | 0.35 | -4.23 | Transport_From | School bus |
| 0.06 | 0.82 | -6.1 | Support_PA | Unsupport | Group 2 (Discovered in DS3,PG1, SubPG1) | | | | | 0.02 | 0.19 | -3.52 | Province | QC |
| 0.11 | 0.84 | -5.1 | GAD7 | moderate | Support | Confidence | SR | Attribute | Attribute Value | 0.03 | 0.35 | -3.3 | Transport_To | School bus |
| 0.29 | 0.86 | -3.5 | Support_Drug_Alc | Unsupport | 0.02 | 0.51 | -42.9 | Smoking | Never | 0.02 | 0.21 | -1.96 | Support_Health | Very Support |
| 0.3 | 0.86 | -2.8 | Support_Tobacco | Unsupport | 0.01 | 0.16 | -29.6 | E-Cigarette | Never | Group 2 (Discovered in DS3,PG2, SubPG1) | | | | |
| 0.17 | 0.86 | -2.4 | Support_Health | Unsupport | 0.01 | 0.39 | -27.6 | Drink | Barely | Support | Confidence | SR | Attribute | Attribute Value |
| Group 2 (Discovered in DS3,PG1, SubPG1) | | | | | 0.01 | 0.27 | -21 | Skip_Class | No | 0.05 | 0.54 | -11.5 | CESD | health |
| Support | Confidence | SR | Attribute | Attribute Value | 0.01 | 0.33 | -13 | Eng_Score | 80%+ | 0.07 | 0.69 | -9.17 | GAD7 | mild |
| 0.12 | 0.78 | -16 | Support_Tobacco | Very Unsupport | 0.01 | 0.3 | -13 | Math_Score | 80%+ | 0.03 | 0.27 | -8.78 | Support_Tobacco | Support |
| 0.13 | 0.8 | -14 | Support_Drug_Alc | Very Unsupport | 0.01 | 0.2 | -11.6 | Education_Like | >=Master | 0.03 | 0.27 | -8.41 | Support_Drug_Alc | Support |
| 0.06 | 0.76 | -14 | Support_Bully | Very Unsupport | 0.01 | 0.17 | -6.61 | Flourish | Flourishing | 0.04 | 0.44 | -6.84 | Support_Bully | Support |
| | | | | | 0.01 | 0.34 | -2.53 | Support_PA | Very Support | 0.05 | 0.54 | -4.26 | PA_Hard | No |
| | | | | | 0.01 | 0.26 | -2.19 | Transport_To | Car(Passenger) | 0.03 | 0.35 | -4.23 | Transport_From | School bus |
| | | | | | | | | | | 0.02 | 0.19 | -3.52 | Province | QC |
| | | | | | | | | | | 0.03 | 0.35 | -3.3 | Transport_To | School bus |
| | | | | | | | | | | 0.05 | 0.51 | -2.21 | Support_Health | Support |

Table 4. Detected Rare Associations for Cannabis Use

Besides the frequent patterns, even some patterns with low frequency or AR are still discovered as significant patterns by PDD. In general, we define a rare pattern as one where its frequency is relatively low in the dataset, but the reason may vary. On one hand, since an outlier must have a low frequency, a rare pattern may be significant, but we may miss it as we consider it as an outlier due to the limitation of the criteria used or class imbalance in the dataset. For example, suppose only a few records are collected for regular cannabis users, which could mean the prevalence is low, but the patterns discovered may still be statistically significant. In this case, we found that even if the frequency of some rare patterns is very low as well as their AR values, PDD still discovers them since they satisfy our criteria: its RAR is greater than the RAR threshold and its AR is less than zero with positive frequency.

Table 4 lists all the rare yet significant patterns or associations discovered by PDD. Some are obviously opposite with frequent patterns, for example, a rare association shows negative associate between Smoking= "Never" and Cannabis Use = "Regular", also the rare association as high academic performance (i.e., Eng_Score= "80%+" and Math_Score= "80%+") and regular cannabis use.

In addition, we also found some interesting, rare patterns, for example, the rare patterns between Cannabis Use = "Non-user" and GAD7= "moderate", which means that a student with moderate GAD7 (scores from 10 to 14) is unlikely to be a non-user. Another rare association is between Cannabis Use = "Non-user" and DERS= "normal" which means that a student with normal DERS (scores from 17 to 30) is unlikely to be a non-user. Another rare pattern is that between transportation and regular cannabis use, which means the student take "School bus" from or to school are unlikely to be a regular cannabis user. These rare cases might be due to other reasons we may not be aware of. PDD findings may lead us to important clues not previously aware of.

**DISCUSSION**

As for data analytics, PDD has a significant advantage over 'black box' machine learning algorithms since it overcomes the major hurdles, namely, interpretability, credibility, and applicability in ML [34]. Firstly, as shown in the randomized controlled trial results, PDD discovers more succinct, interpretable, and hierarchical grouped associations without requiring explicit prior knowledge, even class information. It separated cannabis use groups including a minor "regular cannabis use" group (only 3.11% in the original data). Secondly, association groups obtained from the disentangled spaces are precisely separated corresponding to distinct groups without relying on class information. The detected associations are linked with our designated targets based on their occurrence. The associations can also be retraveled back to the original dataset to identify which records are covered by the discovered associations.

In addition, compared to the ordinary ML methods, PDD can detect interesting patterns from extremely imbalanced datasets directly without using under-sampling or over-sampling strategies. The former might produce results overwhelmed by the majority classes while diminishing significant patterns in minority classes. Finally, significant rare cases are detected by PDD in an interpretable manner. This is very important for healthcare and disease diagnosis since rare patterns may reveal special cases attributed to early detection processes [18].

**LIMITATIONS**

This study has certain limitations. Firstly, only the completed cases of the COMPASS data were used to discover significant frequent/rare associations of behavioral factors in the context of youth cannabis use. Since our objective is to showcase how the PDD algorithm is utilized on population-level health data, any further data analytical approaches such as missing data imputation were not performed in this study. Secondly, only the student-level of the COMPASS data were used. We did not take into consideration the hierarchical and longitudinal design of the COMPASS data to evaluate how changes in the school environment and programs/policies are associated with the initiation and transition of the frequent/rare patterns of youth cannabis use. COMPASS is not designed to be a representative sample and so the discovered associations may not be generalizable to all Canadian youth.

Lastly, from the perspective of the algorithm, in this study, we focus on unsupervised learning, so PDD did not provide prediction results. In addition, since the reasons that caused rare patterns or rare events are various, PDD's results can provide rare or frequent patterns from which are statistically significant but still need experts to interpret the cause(s) of the associations. Hence, as future work, we may consider using discovered patterns by removing the outliers and mislabeled into a separate pool to train a prediction model to produce more accurate results. We will compare our results with those found in literature or by experts who may provide succinct interpretations of these rare patterns and reveal why they occur.

**CONCLUSION**

As a novel pattern discovery method on the imbalanced dataset, PDD renders a set of patterns associated with class/group for explicit and succinct interpretation since it is able to output disentangled patterns concisely and explicitly, which are more specific and distinct for different distinct groups/classes. The results it obtains are statistically robust with comprehensive coverage of succinct, concise, precise, displayable, and less redundant patterns for experts to interpret. PDD also overcomes the problem of imbalanced class [35] . It is also capable of discovering rare yet significant patterns while most methods fail. As a population health data analysis tool on relational data, it has a significant advantage over the 'black box' machine learning algorithms.

The experimental result on the COMPASS dataset --- a large dataset with a highly imbalanced class ratio --- shows that PDD is capable of unveiling patterns for minority targets (Cannabis = Regular). It brings explainable AI to stakeholders and policymakers by providing them with insight from a large amount of data, to enhance their understanding with statistical and rational accountability. Hence, it has a great potential to overcome problems encountered in machine learning and Explainable AI [34] [36]. The evidence from this study is worth observing in the future, namely, the increase, decrease, or maintenance of use patterns, as well as the changing categories of cannabis use and their impacts on youth health outcomes.

In addition, a large number of missing values is a common problem in self-reported or population datasets. In our future work, we will investigate how various missing value imputations will impact PDD performance since the records with missing values may contain significant information.

## DECLARATIONS

**Ethics approval and consent to participate**: The COMPASS study was approved by the University of Waterloo Office of Research Ethics (OR file 17264) and appropriate school boards

**Consent for publication**:  Not applicable.

**Availability of data and materials**: COMPASS data are available for researchers upon successful completion and approval of the COMPASS data usage application (https://uwaterloo.ca/compass-

system/information-researchers). Technical reports detailing COMPASS study methods are available online (https://uwaterloo.ca/compass-system/publications#technical).

**Competing interests**: The authors declare that they have no competing interests.

**Authors' contributions:** PZ and AKCW directed and designed the study, performed the clinical analyses, wrote, and revised manuscript. PZ implemented the algorithm and performed the statistical analyses. YY collected and cleaned the data, wrote, and revise the manuscript. STL is the principal investigator of the COMPASS study, designed the work, revised the manuscript. HC is the principal investigator of this project, designed the work, revised manuscript. KB, ZAB, GM revised manuscript. All authors read and approved the final manuscript.

**Funding:** The COMPASS study has been supported by a bridge grant from the CIHR Institute of Nutrition, Metabolism and Diabetes (INMD) through the "Obesity – Interventions to Prevent or Treat" priority funding awards (OOP-110788; awarded to SL), an operating grant from the CIHR Institute of Population and Public Health (IPPH) (MOP-114875; awarded to SL), a CIHR project grant (PJT-148562; awarded to SL), a CIHR bridge grant (PJT-149092; awarded to KP/SL), a CIHR project grant (PJT-159693; awarded to KP), and by a research funding arrangement with Health Canada (#1617-HQ-000012; contract awarded to SL), and a CIHR-Canadian Centre on Substance Abuse (CCSA) team grant (OF7 B1-PCPEGT 410-10-9633; awarded to SL). The COMPASS-Quebec project additionally benefits from funding from the Ministère de la Santé et des Services sociaux of the province of Québec, and the Direction régionale de santé publique du CIUSSS de la Capitale-Nationale.

**Acknowledgements:** Not applicable.

**Reference**

[1] A. M. Zuckermann, K. Battista, M. de Groh, Y. Jiang and S. T. Leatherdale, "Prelegalisation patterns and trends of cannabis use among Canadian youth: results from the COMPASS prospective cohort study," *BMJ open,* vol. 9, no. 3, p. e026515, 2019.

[2] T. M. Watson and P. G. Erickson, "Cannabis legalization in Canada: how might 'strict'regulation impact youth?," *Drugs: Education, Prevention and Policy,* vol. 26, no. 1, pp. 1-5, 2019.

[3] "Canadian Cannabis Survey 2019 - Summary," Government of Canada, 2019. [Online]. Available: https://www.canada.ca/en/health-canada/services/publications/drugs-health-products/canadian-cannabis-survey-2019-summary.html.

[4] "Canadian Tobacco Alcohol and Drugs (CTADS): 2015 summary," Government of Canada, 2017. [Online]. Available: https://www.canada.ca/en/health-canada/services/canadian-tobacco-alcohol-drugs-survey/2015-summary.html.

[5] "Canadian Tobacco, Alcohol and Drugs Survey, 2015 (updated)," Statistics Canada, 2016. [Online]. Available: https://www150.statcan.gc.ca/n1/daily-quotidien/161109/dq161109b-eng.htm.

[6] "Health Behaviour in School-aged Children: Trends Report 1990-2010," Government of Canada, 25 8 2014. [Online]. Available: https://www.canada.ca/en/public-health/services/health-promotion/childhood-adolescence/programs-initiatives/school-health/health-behaviour-school-aged-children/trends-report-1990-2010.html#a9.0.

[7] A. Butler, "Exploring the association between cannabis use, depression, anxiety and flourishing in youth: a cross-sectional analysis from year 5 of the COMPASS Mental Health pilot data," 2018.

[8] . A. M. Zuckermann, M. R. Gohari, M. de Groh, Y. Jiang and S. T. Leatherdale, "Factors associated with cannabis use change in youth: evidence from the COMPASS study," *Addictive behaviors,* vol. 90, no. Elsevier, pp. 158--163, 2019.

[9] S. Naulaerts, W. Bittremieux, T. Vu, W. Vanden Berghe, B. Goethals and K. Laukens, "A Primer to frequent itemset mining for bioinformatics," *Briefings in bioinformatics,* vol. 16, no. 2, pp. 216-231, 2015.

[10] C. C. Aggarwal and J. Han, Frequent pattern mining, Springer, 2014.

[11] A. K. Wong and Y. Wang, "High-Order Pattern Discovery from Discrete-Valued Data," *IEEE Transaction On Knowledge System,* vol. 9, no. 6, pp. 877-893, 1997.

[12] A. K. Wong and G. C. Li, "Simultaneous pattern and data clustering for pattern cluster analysis," *IEEE Transactions on Knowledge and Data Engineering,* vol. 20, no. 7, pp. 977-923, 2008.

[13] P.-Y. Zhou, G. C. Li and A. K. Wong, "An Effective Pattern Pruning and Summarization Method Retaining High Quality Patterns With High Area Coverage in Relational Datasets," *IEEE Access,* vol. 4, pp. 7847-7858, 2016.

[14] J. Cheng, Y. Ke and W. Ng, "\delta-Tolerance Closed Frequent Itemsets," in *Data Mining, 2006. ICDM'06. Sixth International Conference on. IEEE*, 2006.

[15] P.-Y. Zhou, A. E. Lee, A. Sze-To and A. K. Wong, "Revealing Subtle Functional Subgroups in Class A Scavenger Receptors by Pattern Discovery and Disentanglement of Aligned Pattern Clusters," *Proteomes,* vol. 6, no. 1, p. 10, 2018.

[16] A. K. Wong, A. H. Y. Sze-To and G. L. Johanning, "Pattern to Knowledge: Deep Knowledge-Directed Machine Learning for Residue-Residue Interaction Prediction," *Nature Scientific Reports,* vol. 8, no. 1, pp. 2045-2322, 2018.

[17] P.-Y. Zhou, A. Sze-To and A. K. Wong, "Discovery and disentanglement of aligned residue associations from aligned pattern clusters to reveal subgroup characteristics," *BMC medical genomics,* vol. 11, no. 5, p. 103, 2018.

[18] A. K. Wong, P. Zhou and Z. A. Butt, "Pattern discovery and disentanglement on relational datasets.," *Scientific Reports,* vol. 11, no. 1, pp. 1-11, 2021.

[19] S. Leatherdale, K. Brown, V. Carson, R. Childs, J. A. Dubin, S. J. Elliott, G. Faulkner, D. Hammond, S. Manske, C. M. Sabiston, R. E. Laxer, C. Bredin and A. Thompson-Haile, "The COMPASS study: a longitudinal hierarchical research platform for evaluating natural experiments related to changes in school-level programs, policies and built

environment resources," *BMC Public Health,* vol. 14, no. 331, pp. https://doi.org/10.1186/1471-2458-14-331, 2014.

[20] L. Peng, W. Qing and G. Yujia, "Study on comparison of discretization methods.," in *2009 International Conference on Artificial Intelligence and Computational Intelligence, IEEE*, 2009.

[21] P. Zhou and A. K. Wong, "Explanation and prediction of clinical data with imbalanced class distribution based on pattern discovery and disentanglement," *BMC Medical Informatics and Decision Making,* vol. 21, no. 1, pp. 1-15, 2021.

[22] P. Branco, L. Torgo and R. Ribeiro, "A survey of predictive modelling under imbalanced distributions," *arXiv preprint arXiv:1505.01658,* 2015.

[23] A. Fernández, S. García, M. Galar, R. C. Prati, B. Krawczyk and F. Herrera, Learning from imbalanced data sets, Berlin: Springer, 2018, pp. 1-377.

[24] H. He and Y. Ma, Imbalanced learning: foundations, algorithms, and applications, John Wiley & Sons, 2013.

[25] P. Zhou, A. K. Wong, G. Michalopoulos, R. R. Quinn, M. J. Oliver, Y. Yang, Z. Butt and H. H. Chen, "Revealing Common and Rare Patterns for Peritoneal Dialysis Eligibility Decisions with Association Discovery and Disentanglement," in *2020 IEEE International Conference on Bioinformatics and Biomedicine (BIBM)*, 2020.

[26] Y. S. Koh and N. Rountree, "Finding sporadic rules using apriori-inverse," in *Pacific-Asia Conference on Knowledge Discovery and Data Mining*, Springer, 2005, pp. 97-106.

[27] R. Agrawal and R. Srikant, "Fast algorithms for mining association rules," in *In Proc. 20th int. conf. very large data bases, VLDB*, 1994.

[28] P.-Y. Zhou, A. K. Wong and A. Sze-To, "Discovery and Disentanglement of Protein Aligned Pattern Clusters to Reveal Subtle Functional Subgroups.," in *Bioinformatics and Biomedicine (BIBM), 2017 IEEE International Conference on. IEEE*, Kansas City, MO, USA, 2017.

[29] A. K. Wong, P. Zhou and A. Sze-To, "Discovering Deep Knowledge from Relational Data by Attribute-Value Association," in *Proc. 13th Int. Conf. Data Min. DMIN'17.*, 2017.

[30] A. I. Stiby, M. Hickman, M. R. Munaf, J. Heron, V. L. Yip and J. Macleod, "Adolescent cannabis and tobacco use and educational outcomes at age 16: birth cohort study," *Addiction,* vol. 110, no. 4, pp. 658--668, 2015.

[31] A. Butler, "Exploring the association between cannabis use, depression, anxiety and flourishing in youth: a cross-sectional analysis from year 5 of the COMPASS Mental Health pilot data}," 2018.

[32] S. p. Baggio, A. A. N'Goran, S. p. Deline, J. Studer, M. Dupuis, Y. Henchoz, M. Mohler-Kuo,, J.-B. Daeppen and G. Gmel, "Patterns of cannabis use and prospective associations with health issues among young males," *Addiction,* vol. 109, no. 6, pp. 937--945, 2014.

[33] L. Degenhardt, C. Coffey, H. Romaniuk, W. Swift, J. B. Carlin, W. D. Hall and G. C. Patton, "The persistence of the association between adolescent cannabis use and common mental disorders into young adulthood," *Addiction,* vol. 108, no. 1, pp. 124--133, 2013.

[34] K.-H. Yu, A. L. Beam and I. S. Kohane, "Artificial intelligence in healthcare," *Nature biomedical engineering,* vol. 2, no. 10, pp. 719-731, 2018.

[35] K. Napierala and J. Stefanowski, "Types of minority class examples and their influence on learning classifiers from imbalanced data," *Journal of Intelligent Information Systems,* vol. 46, no. 3, pp. 563-597, 2016.

[36] H. Y. Liang, B. Tsui, H. Xia and etc., "Evaluation and accurate diagnoses of pediatric diseases using artificial intelligence," *Nature Medicine,* vol. 25, pp. 433-438, 2019.